%% file: main.tex
\documentclass[10pt,twocolumn,letterpaper]{article}

\usepackage{cvpr}              %

\input{preamble}

\definecolor{cvprblue}{rgb}{0.21,0.49,0.74}
\usepackage[pagebackref,breaklinks,colorlinks,allcolors=cvprblue]{hyperref}

\usepackage{xspace}

\newcommand{\bemyeyestitle}{%
  \textsc{BeMyEyes}\includegraphics[width=10pt]{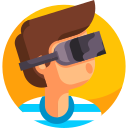}\xspace%
}
\newcommand{\bemyeyes}{%
  \textsc{BeMyEyes}\includegraphics[width=7.5pt]{fig/icon.png}\xspace%
}

\newcommand{\bemyeyeslogo}{%
  \includegraphics[width=7.5pt]{fig/icon.png}\xspace%
}

\newlength\savewidth
\newcommand{\tablestyle}[2]{\setlength{\tabcolsep}{#1}\renewcommand{\arraystretch}{#2}\centering\footnotesize}

\def\paperID{21925} %
\def\confName{CVPR}
\def\confYear{2026}

\title{\includegraphics[width=0.04\textwidth]{fig/icon.png} Be My Eyes: Extending Large Language Models to New Modalities\\Through Multi-Agent Collaboration}

\author{
\textbf{James Y. Huang}\textsuperscript{\rm 1$^*$}~~~
\textbf{Sheng Zhang}\textsuperscript{\rm 2$^*$}~~~
\textbf{Qianchu Liu}\textsuperscript{\rm 2}~~~
\textbf{Guanghui Qin}\textsuperscript{\rm 2}\\
\textbf{Tinghui Zhu}\textsuperscript{\rm 3}~~~
\textbf{Tristan Naumann}\textsuperscript{\rm 2}~~~
\textbf{Muhao Chen}\textsuperscript{\rm 3}~~~
\textbf{Hoifung Poon}\textsuperscript{\rm 2}\\
{\textsuperscript{\rm 1}}University of Southern California~~~
{\textsuperscript{\rm 2}}Microsoft Research~~~
{\textsuperscript{\rm 3}}University of California, Davis\\
\texttt{huangjam@usc.edu}~~~
\texttt{shezhan@microsoft.com}
}

\begin{document}
\maketitle
\def\thefootnote{*}\footnotetext{Equal Contributions.}
\input{sec/0_abstract}    
\input{sec/1_intro}

\input{sec/3_method}
\input{sec/4_experiments}

\input{sec/5_discussion}
\input{sec/2_related}
\input{sec/6_conclusion}

{
    \small
    \bibliographystyle{ieeenat_fullname}
    \bibliography{main}
}

\input{sec/X_suppl}

\end{document}

%% file: preamble.tex
\newcommand{\model}{\mbox{\textsc{BeMyEyes}}\xspace}
\usepackage{tabularx}
\usepackage{soul}
\usepackage{multirow}
\usepackage{makecell}
\newcommand{\hlred}[2][Salmon]{{\sethlcolor{#1} \hl{#2}}}
\newcommand{\hlgreen}[2][YellowGreen]{{\sethlcolor{#1} \hl{#2}}}
\usepackage{tcolorbox}

%% file: sec/0_abstract.tex
\begin{abstract}
Large Language Models (LLMs) have demonstrated remarkable capabilities in challenging, knowledge-intensive reasoning tasks. However, extending LLMs to perceive and reason over a new modality (e.g., vision), often requires costly development of large-scale vision language models (VLMs) with LLMs as backbones. Smaller VLMs are more efficient and adaptable but often lack the broad knowledge and reasoning capabilities of frontier LLMs. In this work, we propose \bemyeyes, a modular, multi-agent framework for extending LLMs to multimodal reasoning by orchestrating collaboration between efficient, adaptable VLMs as perceivers and powerful LLMs as reasoners through conversations. We then introduce a data synthesis and supervised fine-tuning pipeline to train the perceiver agent to effectively collaborate with the reasoner agent. By combining the complementary strengths of perception and reasoning agents, \model avoids the need for training large-scale multimodal models, preserves the generalization and reasoning capabilities of LLMs, and allows flexible extension to new domains and modalities. Experiments show that our framework unlocks the multimodal reasoning capabilities for LLMs, enabling a lightweight and fully open-source solution, i.e. equipping text-only DeepSeek-R1 with Qwen2.5-VL-7B perceiver, to outperform large-scale proprietary VLMs such as GPT-4o on a wide range of knowledge-intensive multimodal tasks. These results demonstrate the effectiveness, modularity, and scalability of our multi-agent approach for building future multimodal reasoning systems.
\end{abstract}

%% file: sec/1_intro.tex
\section{Introduction}

\begin{figure}[!ht]
\centering
\includegraphics[width=0.9\linewidth]{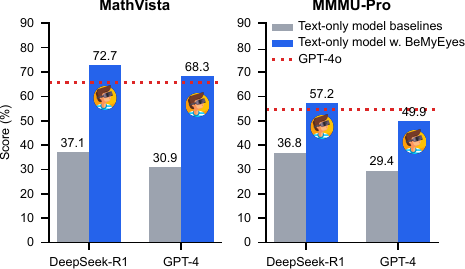} 
\caption{Using \bemyeyes enables text-only models such as DeepSeek-R1 and GPT-4 to reach state-of-the-art performance on challenging multimodal benchmarks \emph{without modifying their parameters}. Grey bars denote text-only baselines, where models receive only the benchmark questions without images. Dotted lines indicate GPT-4o performance.}
\label{fig:hero_bar_plot}
\end{figure}

Large Language Models (LLMs) have demonstrated remarkable capabilities in natural language understanding, reasoning, and generation. Models such as GPT~\citep{openai2024gpt4v}, Qwen~\citep{yang2025qwen3}, and DeepSeek~\citep{guo2025deepseek} illustrate how scaling model size, data, and compute can substantially enhance a model’s knowledge and reasoning abilities. Despite these advances, LLMs remain fundamentally unimodal, operating solely on textual inputs. While they excel at processing and reasoning over language, many real-world tasks require integrating information across multiple modalities, such as vision~\citep{yue2024mmmu, yue2024mmmupro, wang2025muirbench}, video~\citep{li2024mvbench,wu2024longvideobench,fu2025video}, and other sensory inputs. The ability to perceive and reason about non-textual information is crucial for developing general-purpose intelligent systems that can understand and interact with the physical world. Bridging this modality gap has therefore become a central research challenge in extending the capabilities of LLMs toward broader applications.

A dominant approach to overcoming the unimodality of LLMs is to build multimodal language models (MLLMs), such as vision language models (VLMs), that couple pretrained visual encoders with powerful LLM backbones. These multimodal systems~\citep{liu2023visual,lin2024vila,bai2025qwen2,zhu2025internvl3} jointly learn shared representations to enable vision-language alignment and cross-modal reasoning. However, training or adapting such models typically requires substantial computational resources, large-scale multimodal datasets, and often nontrivial architectural modifications when extending to new modalities. In parallel, recent advances in multi-agent systems have demonstrated the potential of agentic, collaborative problem-solving in diverse scenarios such as textual reasoning~\citep{du2023improving}, code generation~\citep{islam2024mapcoder}, and web browsing~\cite{wu2024autogen}. Inspired by these successes, we introduce a multi-agent framework for extending LLMs to new modalities with the help of specialized perceiver agents as the ``eyes'' of LLMs. In this paradigm, an LLM can act as a reasoning agent that leverages its extensive world knowledge and advanced reasoning capabilities, while collaborating with perceiver agents that process and convey information from non-textual inputs. This approach effectively extends LLMs to new modalities without retraining, offering a more scalable and flexible alternative to training large-scale MLLMs.

In this work, we present \bemyeyes, a multi-agent framework designed to extend LLMs to new modalities orchestrating collaboration between efficient, adaptable VLMs as \textit{perceiver agents}, and powerful LLMs as \textit{reasoner agents}. The perceiver agent extracts and describes relevant visual information, while the reasoner agent interprets these descriptions and applies its extensive world knowledge and advanced reasoning capabilities to solve the given task. Through multi-turn conversation, the reasoner can request clarifications or additional visual context, and refine its reasoning based on the perceiver’s responses. This collaboration combines the complementary strengths of small VLMs and large LLMs: the perceiver serves as the ``eyes'' that ground the task in visual evidence, and the reasoner acts as an intelligent expert that drives decision making. To further improve collaboration, we introduce a data synthesis pipeline that allows us to train perceiver agents by distilling strong perceptual and instruction-following capabilities from larger VLMs. As shown in Figure \ref{fig:hero_bar_plot}, \model significantly improves the performance of text-only LLMs on multimodal reasoning, even surpassing large-scale multimodal models, without modifying their parameters.

Our framework provides several key advantages. First, it substantially reduces the cost of developing powerful multimodal models on top of existing LLMs, since only the much more compact perceiver agent needs to be adapted to support new modalities. Second, it preserves the generalization capabilities of LLMs, allowing them to draw on their extensive knowledge and strong reasoning abilities when operating over non-textual inputs. Third, the framework maintains a modular design in which the perceiver and reasoner agents can be swapped independently. As more capable LLMs become available, they can be seamlessly integrated into \model for advanced multimodal reasoning, and lightweight multimodal models that support new modalities can likewise be incorporated to extend task coverage without retraining large multimodal models.
Our main contributions are as follows:
\begin{enumerate}
\item We propose a novel multi-agent framework for extending LLMs to multimodal reasoning by enabling seamless collaboration between specialized perceiver agents and strong reasoner agents via multi-turn conversations.

\item We design a data synthesis pipeline for training perceiver agents to collaborate effectively with reasoner agents on complex multimodal reasoning tasks.

\item We demonstrate the effectiveness and generalizability of our framework across diverse tasks, models, and domains, establishing it as a modular, scalable, and flexible alternative to large-scale multimodal models.

\end{enumerate}

\input{fig/overview}

%% file: fig/overview.tex
\begin{figure*}[ht]
\centering
\includegraphics[width=\textwidth]{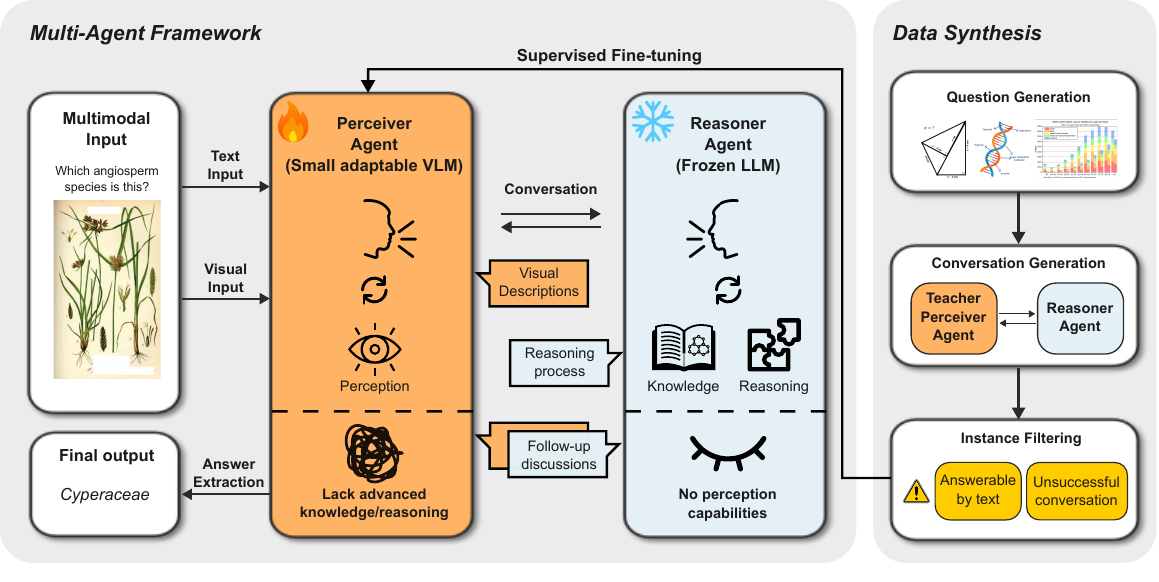} 
\caption{Overview of the \bemyeyes framework. A perceiver agent (small, adaptable VLM) extracts and summarizes visual information from input images, which is then communicated to the reasoner agent (large, frozen LLM) through multi-turn conversations. We also propose a data synthesis and supervised fine-tuning pipeline that allows us to train the perceiver agent to effectively collaborate with the reasoner agent. \model's modular, multi-agent design decouples perception and reasoning, allowing text-only LLMs to perform multimodal reasoning without retraining, while enabling flexible integration of new perceiver or reasoner models.}
\label{fig:overview}
\end{figure*}

%% file: sec/3_method.tex
\section{Method}

\subsection{\bemyeyestitle Overview}
\model is a multi-agent framework designed to extend LLMs to new modalities through collaboration with small, adaptable VLMs acting as perceiver agents (Figure \ref{fig:overview}). The perceiver agent focuses on interpreting and conveying visual information, while the reasoner agent, a large, frozen LLM, leverages its advanced knowledge and reasoning capabilities to solve challenging tasks. This modular design allows LLMs to gain multimodal reasoning abilities without the need for expensive retraining or fine-tuning of large-scale multimodal models. Moreover, as more powerful LLMs emerge, they can be seamlessly integrated into our framework and applied to multimodal tasks, ensuring scalability and adaptability of the system over time.

To enable smooth collaboration between the two agents, we introduce an orchestration mechanism that defines their roles and conversational flow. Furthermore, we propose a data synthesis pipeline and supervised fine-tuning strategy that improve the perceiver’s perception and instruction-following capabilities in our multi-agent setting, leading to more effective collaboration with the reasoner agent.

\subsection{Perceiver Agent}
The primary role of the perceiver agent is to convey visual information to the reasoner agent and to ensure that the reasoner’s outputs remain consistent with the visual evidence present in the input images. Since the perceiver’s responses constitute the sole communication channel through which the reasoner can infer visual context, it is essential that the perceiver agent accurately and comprehensively communicate all relevant visual details. Moreover, the perceiver must actively respond to the reasoner’s requests for specific visual information. To enable such effective collaboration, the perceiver agent is prompted to be aware not only of the necessity of coordination with the reasoner but also of the reasoner’s inherent lack of direct visual perception. This awareness encourages the perceiver to provide detailed, contextually rich descriptions of what it observes.

In \model, perceiver agents are instantiated using small VLMs capable of processing visual inputs. A major advantage of employing small VLMs as perceivers lies in their accessibility and computational efficiency, which facilitates easy development and adaptation of perceiver agents that can effectively collaborate with LLMs on new modalities. While a small VLM alone cannot match the performance of large-scale VLMs, its limitations are effectively compensated by the more capable reasoner agent, which provides advanced reasoning capabilities over the multimodal information supplied by the perceiver.

\subsection{Reasoner Agent}
The reasoner agent is responsible for performing complex reasoning over the visual information provided by the perceiver agent. It is prompted to actively query and retrieve as much relevant visual detail as possible from the perceiver in order to solve the given task. Since the perceiver agent has limited reasoning capabilities, it may omit or overlook certain visual details that are important for accurate problem-solving. To address this limitation, the reasoner agent is instructed to request clarifications or additional descriptions from the perceiver agent when necessary, and to refine its reasoning based on the updated information.

In \model, reasoner agents are instantiated using LLMs that possess strong knowledge and reasoning capabilities but lack native support for visual perception. This makes them ideal candidates for modality extension through multi-agent collaboration. By leveraging the perceiver–reasoner interaction, \model effectively endows the LLM reasoner with multimodal reasoning abilities without requiring costly adaptation of large-scale VLMs.

\subsection{Orchestration}
Both the perceiver and reasoner agents are initialized with system prompts that define their respective roles and the overall problem setting. Each agent is made aware of the complementary expertise of the other: one specializes in perception and the other in reasoning. For a given multimodal reasoning problem with visual and textual inputs, only the perceiver agent initially receives the inputs from both modalities. The two agents then engage in an interactive, multi-turn conversation to solve the problem collaboratively. The reasoner agent typically initiates the dialogue by requesting relevant details about the question, including both visual and textual components. The agents alternate turns, exchanging information and clarifications, until the maximum number of turns is reached. Finally, the perceiver agent is prompted to output the final answer based on the discussions between the two agents.

\subsection{Data Synthesis}
While existing VLMs are generally capable of following regular system prompts, they are not explicitly optimized for our multi-agent setting, where they must coordinate with a separate LLM reasoner. As a result, we observe that they sometimes fail to adhere to their assigned roles, leading to redundant or confusing exchanges between agents~\citep{ossowski2025comma}. In other cases, the perceiver may overlook or misinterpret relevant visual details due to limited perception capabilities when handling complex visual instructions. However, no training data currently exists for fine-tuning perceiver agents in our collaborative multimodal reasoning scenarios. To address this gap and improve the perceiver’s effectiveness in collaborating with the reasoner, we propose a data synthesis pipeline designed to distill the strong perception and instruction-following abilities of larger VLMs into smaller perceiver agents. 

Our data synthesis pipeline consists of three key stages:
\begin{enumerate}
    \item \textbf{Question Generation}:  We begin by collecting a diverse set of images from publicly available datasets. To ensure coverage of knowledge- and reasoning-intensive cases while avoiding overlap with standard evaluation benchmarks, we randomly sample images from CoSyn-400K~\citep{yang2025scaling}, a large collection of computer-generated charts, diagrams, and figures across 10 different categories. We then prompt GPT-4o to generate multimodal reasoning questions that cannot be answered without visual information.
    \item \textbf{Conversation Generation}: For each generated question, we instruct GPT-4o to produce answers under three distinct settings: (1) single-LLM, where only textual input is provided; (2) single-VLM, where both visual and textual inputs are provided; and (3) \model, where GPT-4o is prompted to role-play as both perceiver and reasoner agents, simulating realistic multi-turn conversations between them. We treat the answer from the single-VLM setting as the ground-truth label and sample up to a certain number of conversations from \model, retaining the one that leads to the correct answer. Due to resource constraints, we set the sampling budget to a maximum of eight conversations per question. Preliminary experiments indicate that using GPT-4o to role-play as the teacher perceiver model produces higher-quality data compared to weaker perceiver models such as Qwen2.5-VL, owing to GPT-4o’s superior perception and instruction-following capabilities.
    \item \textbf{Instance Filtering}: We discard instances that either (1) can be answered correctly without images, since these are trivial for multimodal reasoning, or (2) fail to produce any conversations that lead to the correct answer under the \model setting within the sampling budget, typically because the visual information is too complex or abstract to be effectively conveyed through text-based communication. The resulting dataset contains 12,145 multimodal questions, each paired with corresponding images, answers, and simulated conversations between the perceiver and reasoner agents.
\end{enumerate}

\subsection{Supervised Fine-tuning}
Using the synthesized dataset described above, we fine-tune the perceiver agent to improve its ability to collaborate effectively with the reasoner. Specifically, the perceiver is trained on all of its responses within the multi-turn conversations, conditioned on the full prior dialogue history. These responses include: (1) initial descriptions of the visual and textual inputs, (2) follow-up replies to the reasoner’s questions and solution attempts, and (3) the final answers produced after the discussion concludes. The objective is to help the perceiver learn both context-aware visual perception and role-consistent communication within the collaborative reasoning process. We optimize the perceiver using a standard cross-entropy loss over the generated responses:
\[
\mathcal{L}_{\text{SFT}} = - \sum_{t=1}^{T} \log P_{\theta}(y_t \mid y_{<t}, x),
\]
where $x$ denotes the input context (including prior conversation turns and visual inputs), and $y_t$ represents the target token at step $t$. This fine-tuning procedure enables the perceiver to align its perception and communication behaviors with the expectations of the reasoner agent, ultimately leading to more coherent and effective multi-agent collaboration during inference. Note that \model does not require any fine-tuning or architectural modification of the reasoner agent, making the framework both efficient for modality extension and broadly compatible with open-source and proprietary LLMs alike.

%% file: sec/4_experiments.tex
\section{Experiments}

\input{tab/main}

\input{tab/family}

\subsection{Benchmark}
We evaluate our framework on knowledge-intensive multimodal reasoning tasks, which require deep understanding of visual content and advanced reasoning abilities to solve.

\noindent\textbf{MMMU}~\citep{yue2024mmmu} is a comprehensive benchmark for multimodal reasoning covering college-level questions across multiple disciplines. The questions require advanced subject knowledge and complex reasoning to solve.

\noindent\textbf{MMMU Pro}~\citep{yue2024mmmupro} is an expanded and refined version of MMMU, featuring augmented candidate options and filtering out questions that can be answered without visual input. By eliminating questions that can be answered with textual shortcuts, this benchmark emphasizes model’s ability to integrate and reason over both visual and textual information.

\noindent\textbf{MathVista}~\citep{lu2023mathvista} evaluates mathematical reasoning in visual contexts, requiring models to accurately interpret diagrams, charts, and other visual data while applying rigorous mathematical problem-solving skills.

\noindent\textbf{MathVision}~\citep{wang2024measuring} presents a highly challenging set of multimodal math problems sourced from real-world math competitions, pushing models to combine precise visual understanding with advanced mathematical reasoning.

\subsection{Setup}
\paragraph{Models} 
For the perceiver agent, we experiment with two VLM families: Qwen2.5-VL-7B~\citep{bai2025qwen2} and InternVL3-8B~\citep{zhu2025internvl3}. For the reasoner agent, we use GPT-4 (text-only mode of GPT-4o) \citep{openai2024gpt4v} and DeepSeek-R1~\citep{guo2025deepseek}. Although GPT-4o is a multimodal model with native visual perception, for the purpose of reasoner agent in \model we use it as a text-only LLM, providing it only textual inputs during inference. In contrast, DeepSeek-R1 is a strictly text-only model with no visual perception capabilities, making it an ideal candidate for modality extension using our proposed multi-agent approach. This setup allows us to assess how effectively our framework can unlock multimodal reasoning abilities in LLMs that were originally unimodal.

\paragraph{Baselines} 
We compare the performance of \model against two groups of baselines. \textbf{Single-LLM baselines} are restricted to text-only inputs, regardless of whether the underlying model natively supports visual inputs or not. Specifically, all visual content is removed, and only the textual question is provided to the model. \textbf{Single-VLM} baselines are out-of-the-box VLMs that process both textual and visual inputs directly. which represents the standard multimodal approach.

\paragraph{Implementation Details} 
We train the perceiver agent with a learning rate of $5e^{-6}$ over 3 epochs. During evaluation, the temperature is set to 0 to ensure deterministic outputs. The maximum output length per turn is limited to 2048 tokens, and \model is allowed up to 5 interaction turns. For thinking models like DeepSeek-R1, the thinking trace is capped at 4096 tokens using the force-exiting mechanism described in~\citet{muennighoff2025s1}. All models are evaluated under the zero-shot Chain-of-Thought (CoT) setting~\citep{kojima2022large}, and we report accuracy as the primary performance metric across all benchmarks.

\subsection{Main Results}
\paragraph{\model unlocks LLMs' capabilities for multimodal reasoning.} 
As shown in Table \ref{tab:main}, single-LLM baselines generally perform poorly on visual reasoning tasks, highlighting the critical role of visual information. Some questions can still be answered correctly using textual cues alone, but for most knowledge-intensive tasks, access to visual inputs is essential. VLMs such as Qwen2.5-VL and GPT-4o show substantial performance gains when images are available. In contrast, text-only LLMs like DeepSeek-R1 are at a significant disadvantage, despite their superior textual reasoning capabilities compared to other single-LLM baselines. \model unlocks LLMs' multimodal reasoning abilities by collaborating with a small perceiver agent. When paired with a Qwen2.5-VL-7B perceiver, GPT-4 recovers much of the performance of the fully multimodal GPT-4o. Furthermore, \model significatnly boosts the performance of DeepSeek-R1 on multimodal tasks, even outperforming GPT-4o by 0.5\%, 7.1\%, and 12.1\% on MMMU Pro, MathVista, and MathVision, respectively. These results demonstrate that \model effectively extends the reasoning and knowledge capabilities of text-only LLMs to multimodal tasks.

\input{tab/med}

\input{tab/ablation}

\paragraph{\model brings consistent improvements across model families.} 
To evaluate the robustness of our framework, we train and test \model using the same dataset and benchmarks but replace the perceiver agent with InternVL3-8B. As reported in Table \ref{tab:family}, all previous observations are consistent: \model continues to provide substantial performance gains for LLMs across all benchmarks. In particular, using InternVL3-8B as ``eyes'', DeepSeek-R1 consistently outperforms GPT-4o on all multimodal reasoning tasks, demonstrating that our approach generalizes effectively across different model families and is not limited to a specific model pairing.

\paragraph{\model generalizes well to specialized domains.} 
To evaluate the generalizability of \model to domain-specific applications, we test it on MMMU Med~\citep{yue2024mmmu} and MMMU Pro Med~\citep{yue2024mmmupro}. These benchmarks require both precise perception of medical imagery and advanced domain-specific knowledge reasoning. For the perceiver agent, we use Lingshu-7B~\citep{xu2025lingshu}, a variant of Qwen2.5-VL-7B specializing in multimodal medical reasoning. We fine-tune this perceiver using the same synthetic data used for Qwen2.5-VL, without incorporating any medical-specific training data. As shown in Table \ref{tab:med}, \model maintains strong performance on par with GPT-4o on challenging medical reasoning tasks, demonstrating its ability to generalize effectively to specialized domains. These results suggest that \model provides a scalable and efficient alternative for extending LLMs to new domains without requiring large-scale domain-specific multimodal training.

\subsection{Ablations}

\paragraph{Supervised Fine-tuning promotes effective collaboration.} 
To evaluate the impact of supervised fine-tuning (SFT) on the performance of \model, we conduct an ablation study comparing (1) a non-SFT version of \model, where the perceiver agent is used without any additional training, and (2) the fine-tuned perceiver agent evaluated as a standalone VLM. As shown in Table \ref{tab:ablation}, applying supervised fine-tuning consistently improves the performance of \model, highlighting the crucial role of learning effective collaboration. Interestingly, this performance gain cannot be attributed merely to improvements in the perceiver’s individual reasoning ability. When evaluated independently as a VLM, the perceiver shows minimal or no improvement across most benchmarks after fine-tuning. This suggests that supervised fine-tuning primarily enhances the perceiver’s ability to collaborate with the reasoner, refining its perception and communication, rather than simply improving its standalone reasoning capacity.

\paragraph{Multi-turn conversation benefits collaborative problem solving.} 
We further conduct an ablation study to examine the effect of multi-turn interaction between agents. Specifically, we evaluate \model in a single-turn setting, where the system prompts and the orchestration process are slightly modified to restrict the conversation to one turn. In this setting, the perceiver agent is allowed to respond only once, communicating all relevant visual information to the reasoner before the latter produces its final answer. This configuration removes the possibility of clarification, feedback, or iterative refinement during reasoning. While most questions can still be answered correctly within a single turn thanks to the strong reasoning capabilities of the LLM reasoner, a performance gap remains between single-turn and multi-turn \model. This suggests that iterative communication plays an important role in improving the overall reasoning process.

\input{tab/case}

%% file: tab/main.tex
\begin{table*}[t]
\centering
\begin{tabular}{lcccc}
\toprule
\textbf{Model} & \textbf{MMMU}& \textbf{MMMU Pro}& \textbf{MathVista}& \textbf{MathVision}\\
\midrule
\multicolumn{5}{c}{\textit{Single-LLM Baselines}} \\
\midrule
Qwen2.5-VL-7B (text-only) & 40.7 & 22.8 & 30.6 & 25.4 \\
GPT-4 & 48.9 & 29.4 & 30.9 & 28.8 \\
DeepSeek-R1 & 54.8 & 36.8 & 37.1 & 42.6 \\
\midrule
\multicolumn{5}{c}{\textit{Single-VLM Baselines}} \\
\midrule
Qwen2.5-VL-7B & 54.0 & 39.8 & 65.1 & 27.4 \\
GPT-4o & \textbf{68.3} & 56.7 & 65.6 & 36.4 \\
\midrule
\multicolumn{5}{c}{\textit{Our Methods (using Qwen2.5-VL-7B as eyes)}} \\
\midrule
GPT-4 + \bemyeyes & 64.6 & 49.9 & 68.3 & 32.8 \\
DeepSeek-R1 + \bemyeyes & 67.4 & \textbf{57.2} & \textbf{72.7} & \textbf{48.5} \\
\bottomrule
\end{tabular}
\caption{Performance of \model when using Qwen2.5-VL-7B as the perceiver agent on four multimodal reasoning benchmarks. The framework consistently improves multimodal reasoning, with DeepSeek-R1 outperforming large-scale multimodal models such as GPT-4o, demonstrating the effectiveness of \model in extending LLM reasoning capabilities to new modalities.}
\label{tab:main}
\end{table*}

%% file: tab/family.tex
\begin{table}[t]
\footnotesize
\centering
\tablestyle{1pt}{1}
\begin{tabular}{lcccc}
\toprule
\multirow{2}{*}{\textbf{Model}} & \multirow{2}{*}{\textbf{MMMU}} & \textbf{MMMU} & \multirow{2}{*}{\textbf{MathVista}} & \multirow{2}{*}{\textbf{MathVision}}\\
& & \textbf{Pro} & & \\
\midrule
\multicolumn{5}{c}{\textit{Baselines}} \\
\midrule
InternVL3-8B (text-only) & 47.1 & 25.7 & 35.0 & 26.6 \\
InternVL3-8B & 60.1 & 44.5 & 64.7 & 29.2 \\
GPT-4o & 68.3 & 56.7 & 65.6 & 36.4 \\
\midrule
\multicolumn{5}{c}{\textit{Our Methods (using InternVL3-8B as eyes)}} \\
\midrule
GPT-4 + \bemyeyeslogo & 62.7 & 49.5 & 70.7 & 32.9 \\
DeepSeek-R1 + \bemyeyeslogo & \textbf{69.7} & \textbf{58.5} & \textbf{73.1} & \textbf{50.6} \\
\bottomrule
\end{tabular}
\caption{Performance of \model when using InternVL3-8B as the perceiver agent across four multimodal reasoning benchmarks. \model consistently improves multimodal reasoning across all tasks, demonstrating that \model is robust across different model pairings.}
\label{tab:family}
\end{table}

%% file: tab/med.tex
\begin{table}[t]
\centering
\setlength{\tabcolsep}{4pt}
\footnotesize
\begin{tabular}{lcc}
\toprule
\multirow{2}{*}{\textbf{Model}} & \textbf{MMMU} & \textbf{MMMU Pro} \\
& \textbf{Med} & \textbf{Med} \\
\midrule
\multicolumn{3}{c}{\textit{Baselines}} \\
\midrule
GPT-4 & 55.3 & 32.2 \\
DeepSeek-R1 & 56.0 & 36.0 \\
Lingshu-7B & 65.3 & 35.0 \\
GPT-4o & \textbf{78.0} & \textbf{59.1}\\
\midrule
\multicolumn{3}{c}{\textit{Our Methods (using Lingshu-7B as eyes)}} \\
\midrule
GPT-4 + \bemyeyes & 74.0 & 50.4 \\
DeepSeek-R1 + \bemyeyes & 76.7 & 58.7 \\
\bottomrule
\end{tabular}
\caption{Performance of \model on multimodal medical reasoning tasks when using Lingshu-7B as the perceiver agent.}
\label{tab:med}
\end{table}

%% file: tab/ablation.tex
\begin{table}[t]
\footnotesize
\centering
\setlength{\tabcolsep}{2pt}
\begin{tabular}{lcccc}
\toprule
\multirow{2}{*}{\textbf{Model}} & \multirow{2}{*}{\textbf{MMMU}} & \textbf{MMMU} & \multirow{2}{*}{\textbf{MathVista}} & \multirow{2}{*}{\textbf{MathVision}}\\
& & \textbf{Pro} & & \\
\midrule
Qwen2.5-VL-7B & 54.0 & 39.8 & 65.1 & 27.4 \\
Qwen2.5-VL-7B-SFT & 54.6 & 40.6 & 67.7 & 26.8 \\
\midrule
\model (no SFT) & 62.7 & 49.1 & 65.1 & 30.6 \\
\model (single turn) & 64.0 & 48.3 & 67.2 & 31.0 \\
\model & \textbf{64.6} & \textbf{49.9} & \textbf{68.3} & \textbf{32.8} \\
\bottomrule
\end{tabular}
\caption{Ablation results for \model. Performance drops are observed when removing supervised fine-tuning (SFT) or restricting the framework to single-turn conversations, demonstrating that both fine-tuning and multi-turn communication are crucial for effective collaboration and robust multimodal reasoning.}
\label{tab:ablation}
\end{table}

%% file: tab/case.tex
\begin{table*}[t]
\scriptsize
\centering
\begin{tabularx}{\textwidth}{l|l|X}
\toprule
\textbf{Question} & \textbf{Model} & \textbf{Response} \\
\midrule
\multirowcell{3}{Which angiosperm species is this?\\ \\ \includegraphics[width=0.17\textwidth]{fig/test_Biology_160_1.jpg}} & Qwen2.5-VL-7B & To determine the correct answer, let's analyze the characteristics of the plant shown in the image: 1. The plant has long, narrow leaves, which are typical of grasses ... \hlred{Given these characteristics, the plant appears to be a type of grass.} Now, let's match this with the given options: \hlred{A. Poaceae - This is the family of grasses} ... \hlred{G. Cyperaceae - This is the family of sedge plants} ... Based on the characteristics of the plant, the correct family is Poaceae. Answer: A\\
\cmidrule{2-3}
& GPT-4o & To identify the angiosperm species in the illustration, let's analyze the characteristics of the plant depicted: 1. **Leaf Structure** ... 3. **Growth Habit**: The plant appears to have a tufted growth habit with a basal cluster of leaves, \hlgreen{which is characteristic of sedges (Cyperaceae) rather than true grasses (Poaceae).} ... Based on these observations, the plant in the illustration is most likely from the Cyperaceae family, which includes sedges. Answer: G\\
\cmidrule{2-3}
& \bemyeyes & \textbf{Perceiver}: The question is ... Description of the image: The image is a botanical illustration of a plant species. It shows several parts of the plant, including leaves, flowers, and roots. The leaves are long and narrow, resembling grass blades ... \hl{The overall appearance suggests a species that is likely aquatic or semi-aquatic, given the structure of the leaves and flowers.} Based on this description, could you help identify the angiosperm species?\\ \\
& & \textbf{Reasoner}: Thank you for the detailed description. Let's analyze the information to identify the angiosperm species: ... Given these characteristics, let's consider the options: ... Based on the description, the plant seems to fit well with either Poaceae or Cyperaceae due to the leaf shape and flower structure. \hlgreen{However, the aquatic or semi-aquatic hint leans more towards Cyperaceae, as sedges are often found in such environments.} Therefore, the most likely answer is **G. Cyperaceae**.\\
\bottomrule
\end{tabularx}
\caption{Responses generated by baselines and \model on an example from MMMU Pro. Qwen2.5-VL-7B by itself fails to reach the correct answer due to \hlred{reasoning errors}, while text-only GPT-4 reasoner in \model is able to recover the \hlgreen{correct reasoning traces} based on the \hl{visual clues} provided by the perceiver agent.}
\label{tab:case}
\end{table*}

%% file: sec/5_discussion.tex
\section{Discussion}

\input{fig/breakdown}

\subsection{Error Breakdown}
To better understand the individual contributions of each agent in \model, we conduct an error breakdown analysis using examples from MMMU Pro. Each example is labeled based on the correctness of answers under three settings: (1) single-perceiver (Qwen2.5-VL-7B), (2) single-reasoner-with-vision (GPT-4o), and (3) \model. This comparison provides a clearer view of how the perceiver and reasoner perform both individually and collaboratively, offering insights into the interplay between perception and reasoning in the multi-agent framework. As illustrated in Figure \ref{fig:breakdown}, we group all examples based on answer correctness across the three settings and order the groups based on their size.

From this breakdown, we observe that more than half of the examples fall into the first two groups, representing cases that are either trivially easy or universally challenging, where all or none of the settings yield correct answers. The third-largest group includes cases where the perceiver alone would fail, but the reasoner successfully captures visual information conveyed by the perceiver and recovers the correct answer, demonstrating successful collaboration. Conversely, the fourth-largest group captures instances where the fully multimodal reasoner would answer correctly in isolation, but is misled when reasoning jointly with the perceiver, often due to perception or communication errors. Interestingly, a small portion of examples show disagreement between single-model and multi-agent predictions, where \model’s answer contradicts the consensus of its individual models. These cases suggest behavioral inconsistencies introduced by multi-agent dynamics, both beneficial and detrimental, consistent with prior observations that multi-agent collaboration can improve reasoning or introduce new sources of error \citep{du2023improving, wynn2025talk}.

\subsection{Case Study}
In Table \ref{tab:case}, we showcase the responses generated by baselines and \model on a typical example from MMMU Pro. Qwen2.5-VL-7B fails to correctly identify the species shown in the image due to limited domain knowledge and reasoning capability, while GPT-4o can answer the question correctly. When GPT-4 is used as the reasoner agent within \model, it can still infer the correct answer by leveraging the perceiver’s detailed visual descriptions. This demonstrates how effective communication and role specialization in \model enable the reasoner to compensate for its lack of direct visual perception through collaboration, achieving performance comparable to a fully multimodal model.

\subsection{Limitations and Future Work}
While \model demonstrates strong performance in efficient and modular modality extension for LLMs, several open questions remain for future exploration. First, our experiments focus exclusively on vision as the target modality, although the framework is general and can naturally be extended to other modalities such as audio and video. Second, a comparison against an upper bound established by a hypothetical, fully-trained multimodal DeepSeek-R1 would provide deeper insights into the effectiveness of our approach; however, developing such large-scale models remains beyond our current scope. Finally, given the recent success of reinforcement learning (RL) in improving multimodal reasoning~\citep{li2025self}, incorporating an RL-based training pipeline may further improve the effectiveness of \model, an exciting direction we leave for future work.

%% file: fig/breakdown.tex
\begin{figure}[t]
\centering
\includegraphics[width=\linewidth]{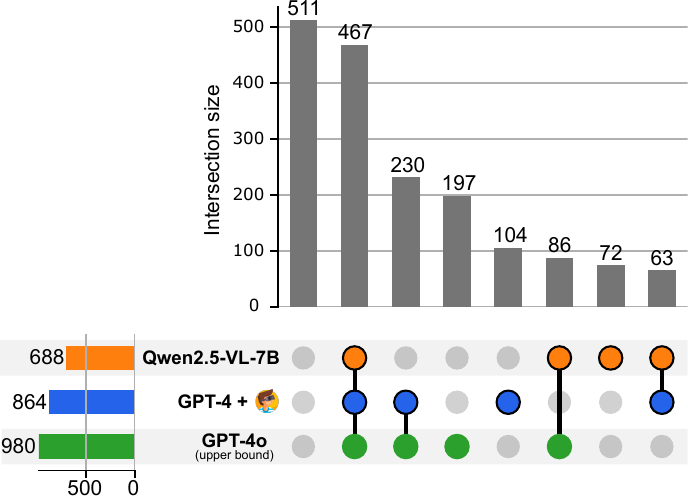} 
\caption{Error breakdown on MMMU Pro. We group all examples based on answer correctness under the single-perceiver, single-reasoner-with-vision, and \model settings, respectively. The groups are represented by a three-light code signaling correct answers in each setting, and we order the groups based on their size. The bottom-left bar chart shows the total number of correctly answered examples for each setting.}
\label{fig:breakdown}
\end{figure}

%% file: sec/2_related.tex
\section{Related Works}

\paragraph{Vision Language Models}
Large vision language models (VLMs) have received considerable attention for their ability to process and reason across visual and textual modalities, enabling a wide range of applications. Recent approaches~\citep{alayrac2022flamingo,li2022blip,openai2024gpt4v} have focused on aligning visual and language representations through instruction fine-tuning. For example, BLIP-2~\citep{li2023blip} and LLaVA~\citep{liu2023visual} equip base LLMs with image encoders, while more recent works extend this paradigm to video inputs~\citep{zhang2023video}, further enriching the perceptual context available to MLLMs.
Despite these advances, existing VLMs face notable limitations. They are often prone to knowledge conflicts~\citep{zhu2024unraveling}, where visual reasoning may contradict prior textual knowledge, and exhibit language biases~\citep{zhang2024debiasing,wang2024mdpo}, which can degrade reasoning performance. To mitigate language biases, prior works have proposed to disentangle visual perception and language reasoning~\cite{jia2024describe,xia2025visionary,li2025self}, adopting a caption-then-reason paradigm for robust multimodal reasoning with a single VLM. However, none of these works explore the collaboration of VLMs and LLMs for modality extension in a multi-agent setting.

\paragraph{Modality Extension}
Modality extension is a key research area aimed at transforming LLMs, primarily designed for textual processing, into multimodal language models (MLLMs) capable of integrating and reasoning over additional modalities, such as vision~\citep{yue2024mmmu, yue2024mmmupro, wang2025muirbench} and video~\citep{li2024mvbench,wu2024longvideobench,fu2025video}. The dominant approach involves modality fine-tuning of a pre-trained LLM \citep{liu2023visual,li2023blip,lin2024vila,bai2025qwen2,zhu2025internvl3}, leveraging paired language–modality data along with modality-specific instruction datasets. This process serves two main objectives: (1) aligning the representations of the new modality with the pre-existing language space, and (2) enabling the MLLM to follow modality-specific instructions effectively~\citep{liu2023visual}. Despite its popularity, modality fine-tuning faces several notable limitations. MLLMs often demonstrate reduced reasoning capabilities compared to their text-only counterparts and can be more vulnerable to adversarial or malicious inputs~\citep{zhu2025extending}. Furthermore, the success of this approach critically depends on the availability of large-scale, high-quality modality-aligned datasets, which are costly and labor-intensive to construct. These challenges motivate alternative approaches for extending LLMs to new modalities without full-scale retraining.

\paragraph{Multi-Agent Collaboration} 
Multi-agent collaboration enables multiple agents to work together to solve complex tasks that are difficult for a single agent to handle~\citep{islam2024mapcoder,guo2024large,fourney2024magentic}. Previous works have demonstrated the benefit of collaborative reasoning in a unimodal, multi-agent setting where discussion between agents offers diverse perspectives and more robust solutions~\citep{chan2024chateval,chen2024reconcile,liang2024encouraging,khan2024debating,eisenstein2025don}. Inspired by these successes, we believe multi-agent collaboration provides a alternative framework for overcoming the limitations of unimodal LLMs and extending them to new modalities by collaborating with specialized perceiver agents, while preserving the reasoning capabilities of LLMs and avoiding the cost of training large-scale MLLMs.

%% file: sec/6_conclusion.tex
\section{Conclusion}
In this work, we introduced \model, a modular multi-agent framework that extends large language models (LLMs) to new modalities through collaboration with small, efficient vision-language models (VLMs). Comprehensive experiments on multiple multimodal reasoning benchmarks demonstrate that \model substantially improves the multimodal reasoning performance of text-only LLMs, achieving results that rival or surpass large-scale VLMs such as GPT-4o. These findings highlight the potential of multi-agent collaboration as a scalable and flexible alternative to large-scale multimodal models.

%% file: sec/X_suppl.tex
\clearpage
\appendix
\onecolumn

\section{Prompt}

\begin{tcolorbox}[colback=white, colframe=gray!80!gray, boxrule=1pt, title=Prompt for Single-Model Baselines]
\texttt{Answer the preceding multiple choice question. The last line of your response should be of the following format: ``Answer: \$LETTER'' (without quotes) where LETTER is one of the options. Think step by step before answering.}
\end{tcolorbox}

\begin{tcolorbox}[colback=white, colframe=gray!80!gray, boxrule=1pt, title=System Prompt for the Perceiver Agent]
\texttt{Your task is to answer a given multiple choice question about an image with the help of the expert. The expert does not have access to the question, the options, or the image, so you should state the exact question and the options, and provide a detailed description of the image to the expert.}
\end{tcolorbox}

\begin{tcolorbox}[colback=white, colframe=gray!80!gray, boxrule=1pt, title=System Prompt for the Reasoner Agent]
\texttt{Your task is to help the client answer a multiple choice question about an image. Only the client have access to the question, the options, and the image, so you should try to gather from the client as much information as needed to answer the question. Make sure you fully understand the question and verify details about the image that may be relevant to each option before answering the question.}
\end{tcolorbox}

\begin{tcolorbox}[colback=white, colframe=gray!80!gray, boxrule=1pt, title=Prompt for Question and Image Descriptions]
\texttt{Hi, I'm the expert here. I heard you have a multiple choice question about an image and I can help you with that. Could you state the exact question, the options, and provide a detailed description of the image?}
\end{tcolorbox}

\begin{tcolorbox}[colback=white, colframe=gray!80!gray, boxrule=1pt, title=Prompt for Answer Extraction]
\texttt{Now it's time to write the final answer. Your response should be of the following format: ``Answer: \$LETTER'' (without quotes) where LETTER is one of the options.}
\end{tcolorbox}